%% file: sn-article.tex
\documentclass[pdflatex, sn-vancouver]{sn-jnl}
\usepackage{fix-cm}
\usepackage[T1]{fontenc}
\usepackage{natbib}
\usepackage{mathrsfs}

\jyear{2021}%

\theoremstyle{thmstyleone}%
\graphicspath{{./figures/}}
\theoremstyle{thmstyletwo}%
\theoremstyle{thmstylethree}%
\begin{document}

\title[ResNLS: An Improved Model for Stock Price Forecasting]{ResNLS: An Improved Model for Stock Price Forecasting}

\author[1]{\fnm{Yuanzhe} \sur{Jia}}\email{yjia5612@uni.sydney.edu.au}
\author[1]{\fnm{Ali} \sur{Anaissi}}\email{ali.anaissi@sydney.edu.au}
\author[1,2]{\fnm{Basem} \sur{Suleiman}}\email{basem.suleiman@sydney.edu.au}

\affil[1]{\orgname{University of Sydney}, \orgaddress{\city{Sydney}, \state{NSW}, \country{Australia}}}
\affil[2]{\orgname{University of New South Wales}, \orgaddress{\city{Sydney}, \state{NSW}, \country{Australia}}}

\abstract{
Stock prices forecasting has always been a challenging task. 
Although many research projects try to address the problem, few of them pay attention to the varying degrees of dependencies between stock prices. 
In this paper, we introduce a hybrid model that improves the prediction of stock prices by emphasizing the dependencies between adjacent stock prices.
The proposed model, ResNLS, is mainly composed of two neural architectures, ResNet and LSTM. ResNet serves as a feature extractor to identify dependencies between stock prices, while LSTM analyzes the initial time series data with the combination of dependencies, which are considered as residuals. 
Our experiment reveals that when the closing price data for the previous 5 consecutive trading days is used as input, the performance of the model (ResNLS-5) is optimal compared to those with other inputs. 
Furthermore, ResNLS-5 demonstrates at least a 20\% improvement over current state-of-the-art baselines.
To verify whether ResNLS-5 can help clients effectively avoid risks and earn profits in the stock market, we construct a quantitative trading framework for back testing. The result shows that the trading strategy based on ResNLS-5 predictions can successfully mitigate losses during declining stock prices and generate profits in periods of rising stock prices.
The relevant code is publicly available on GitHub\footnote{\url{https://github.com/yuanzhe-jia/stock-price-forecasting}}.
}


\maketitle
\input{ae_intro.tex}

\input{ae_relatedWork.tex}

\input{ae_background.tex}

\input{ae_method.tex}

\input{ae_results.tex}

\input{ae_conclusion.tex}

\bibliography{ae_ref}
\end{document}

%% file: ae_intro.tex
\subsection{Introduction}

The exploration of stocks has never stopped since the inception of the stock market, and stock price forecasting stands out as the most compelling and widely discussed topic. However, many factors affect stock price fluctuations. From a macro perspective, stock prices will be influenced by several factors including international environments, national policies, economic development levels and social hot issues \cite{shiller1984stock, conover1999monetary, pastor2012uncertainty, huy2021impacts}. From a micro point of view, stock prices are also affected by factors such as the company operations, competition, disclosure, and investor sentiment \cite{iqbal2000stock, fisher2000investor, ozoguz2013information}. These factors interact, making the internal mechanism of the stock market difficult to understand, and the change cycle even more difficult to grasp \cite{king1966market, liu2015economic}.

The rise of AI technology makes it increasingly possible to predict future \cite{arslanian2019future, strader2020machine, chopra2021application,anaissi2021intelligent, anaissi2021online}, helping us find inspiration and motivation to study the problem of stock price forecasting. However, the mainstream approaches, like CNN-based, RNN-based and Transformer-based prediction models, 
do not emphasize distinguishing the degree of dependencies between stock prices, especially between adjacent price data and data far apart. Therefore, our key goal is to propose a time-series prediction model that pays much attention to adjacent data characteristics.
The contributions of this paper are as follows: (a) analysing the limitations of mainstream prediction models for stock price forecasting; (b) proposing a sequential neural network that values the features of different dependencies between stock prices and verify its superiority; (c) back testing the model and demonstrating that it can help to avoid risks and make profits for clients.

This paper is organized as follows. Section \ref{sec:relatedwork} is about related work, mainly discusses and analyses the contributions and limitations of the mainstream stock price prediction approaches in the last 20 years. Section \ref{sec:prelim} introduces the preliminaries required for understanding the model we proposed in this paper. In section \ref{sec:method} we describe our research ideas and model structure in detail. In section \ref{sec:experiment} we clarify the dataset used for experiments, the model training procedure, the evaluation metrics, as well as how to conduct experiments on exploratory data analysis, model comparison and back testing to verify the predictive ability, stability and applicability of our model. Finally,  we conclude the paper in section \ref{sec:conclusion} and look to the future.

%% file: ae_relatedWork.tex
\section{Related Work}
\label{sec:relatedwork}

Stock price forecasting has earned a lot of interests in recent years and has attracted many researchers working in academia and industry. Although the EMH \cite{fama1965behavior} indicates that it is impossible to predict future prices based on historical data, many researchers have attempted to demonstrate the validity of price prediction \cite{smith2003constructivist, bisen2013testing, lekovic2018evidence, spulbar2021critical, stephens2021new}. In recent years, scholars have pinned their hopes on AI technology to improve the accuracy of the prediction models. Thus, conference papers, journal articles and academic reports related to this area in the past 20 years have been analyzed. We focus on the work that proposed fruitful new approaches and group them into four categories, which are SVM-based models, tree-based models, neural network models and hybrid neural network models. Since this section contains a large number of technical terms, the notations are summarized in Table \ref{tab:notations} for the convenience.

\begin{table}
\caption{Notations used in this paper.}
\label{tab:notations}
\centering
\begin{tabular}{ p{2cm} p{9cm} }
\hline
Abbreviation & Description \\
\hline
AI & Artificial Intelligence \\
ANFIS & Adaptive Neuro Fuzzy Inference System \\
ANN & Artificial Neural Network \\
ARIMA & Auto-regressive Integrated Moving Average \\
BiLSTM & Bi-directional Long Short-term Memory \\
BSO & Brain Storm Optimization \\
CNN & Convolution Neural Network \\
CSI 300 & The price-weighted average index of 300 top-rated listed companies on the Shanghai Stock Exchange and Shenzhen Stock Exchange \\
EMH & Efficient Market Hypothesis \\
LSTM & Long Short-term Memory \\
NASDAQ 100 & The price-weighted average index of the top 100 largest non-financial companies listed on the NASDAQ Exchange \\
NIKKEI 225 & The price-weighted average index of 225 top-rated Japanese companies listed in the First Section of the Tokyo Stock Exchange \\
NLP & Natural Language Processing \\
PCA & Principal Component Analysis \\
PSO & Particle Swarm Optimization \\
ResNet & Residual Neural Network \\
ResNLS & The model proposed in this research \\
RNN & Recurrent Neural Network \\
S\&P 500 & The price-weighted average index of the top 500 most influential manufacturing companies compiled by Standard \& Poor's \\
SSE & The price-weighted average index covering all listed companies on the Shanghai Stock Exchange \\
SVM & Support Vector Machine \\
SVR & Support Vector Regression \\
\hline
\end{tabular}
\end{table}

\textbf{SVM-based models and tree-based models}. SVM was earlier used for stock price prediction \cite{cao2001financial, kim2003financial, huang2005forecasting, gavrishchaka2006support, yeh2011multiple}. Huang, W. et al. \cite{huang2005forecasting} adopted SVM to predict the NIKKEI 225 Index. However, the authors stated at the end that in view of the vulnerability of SVM to the over-fitting problem, future work should consider combining it with other algorithms. Inspired by the idea, Wang, J. et al. \cite{wang2016improved} developed a hybrid SVR model by combining PCA and BSO algorithms. Oztekin, A. et al. \cite{oztekin2016data} developed an ensemble model which integrates ANFIS and SVM; Xiao, C. et.al.  \cite{xiao2020stock} combined the ARIMA with SVM. Those hybrid models take into account the linear and nonlinear characteristics of stock fluctuations, so as to achieve higher prediction performance. Tree-based models also have good applications in stock price forecasting. Wu, M.-C. et al. \cite{wu2006effective} used a combination of Decision Tree and Filter Rule to predict the NASDAQ 100 Index. Filter Rule is a common way used by traditional investors to analyse the timing of stock buying and selling. Combined with Decision Tree, it can be considered as a supplement of empirical analysis to machine intelligence. However, a common problem faced by these models is the low prediction accuracy.

\textbf{Neural network models}. With the popularity of deep learning, prediction models based on neural networks emerge in an endless way. Many works have been done using ANN for stock price prediction \cite{qi2008trend, yu2009neural, mostafa2010forecasting, guresen2011using, qiu2016application} and achieved better performance than SVM-based and tree-based models. Some researchers took different approaches on training data. In the prediction task of the S\&P 500 Index, Chen, Y.-C., \& Huang, W.-C. \cite{chen2021constructing} trained a CNN model with different input features. Their research proved that using the gold price, the gold volatility index, the crude oil price and the crude oil price volatility index as input features allowed deep models to accurately predict future stock prices. Other studies focused on modifying the vanilla neural networks for better predictions. For example, Chandriah, K.-K., \& Naraganahalli, R.-V. \cite{chandriah2021rnn} found that using the modified-Adam algorithm to optimize parameters in RNN can enhance the predictive ability of the network and make it suitable for long-time prediction tasks. Lv, L. et al. \cite{lv2018improved} used the PSO algorithm to optimize the hidden states of the vanilla LSTM. Wang, X. et al. \cite{wang2021forecasting} introduced a 1-arctan function to improve the forget gate of the vanilla LSTM. All the studies achieved better performance in accuracy, reliability and adaptability.

\textbf{Hybrid neural network models}. In recent years, hybrid neural networks have gradually occupied the mainstream of stock price prediction tasks. The core idea is to use the encoder-decoder framework, first apply the encoder to extract features from the original input, and then feed the extracted features into the decoder for prediction. Some researchers \cite{kim2018forecasting} combined the vanilla LSTM with various generalized auto-regressive conditional heteroscedasticity-type models and achieved excellent prediction results. Another representative work \cite{yang2020html} proposed a novel hierarchical transformer with the multi-task architecture designed to harness the text and audio data from quarterly earnings conference calls to predict future stock prices in the short and long term. A recent study \cite{ko2021lstm} used BERT, which has been shining in the NLP field, as an encoder to perform sentiment analysis on stock-related comments in social media, then integrated the output features with initial stock prices to enhance LSTM performance. The prediction accuracy of the model is 12.05\% higher than that of the vanilla LSTM.

In conclusion, neural network-based models are now absolutely dominant when it comes to stock price forecasting. However, most of them overlook the differences in the degree of dependencies between stock prices, let alone consider both the dependency differences and time-series characteristics simultaneously. To address this limitation, our paper introduces a hybrid model that balances both aspects effectively.

%% file: ae_background.tex
\section{Preliminary}
\label{sec:prelim}

Recurrent neural networks \cite{elman1990finding} are commonly used to process the time-series data like stock prices. In RNN, each neuron has two kinds of input data, one is the hidden state from the previous neuron, the other is the input data from the current time step. The main advantage of RNN is the memory cell (i.e. the hidden state) that can be updated at each time step. So the last hidden state will theoretically contain all the useful information extracted from the sequential input after training. But RNN is not that robust as the gradients flowing through the network are sometimes too large or too small.

\begin{figure}[t]
\centerline{\includegraphics[width=22pc]{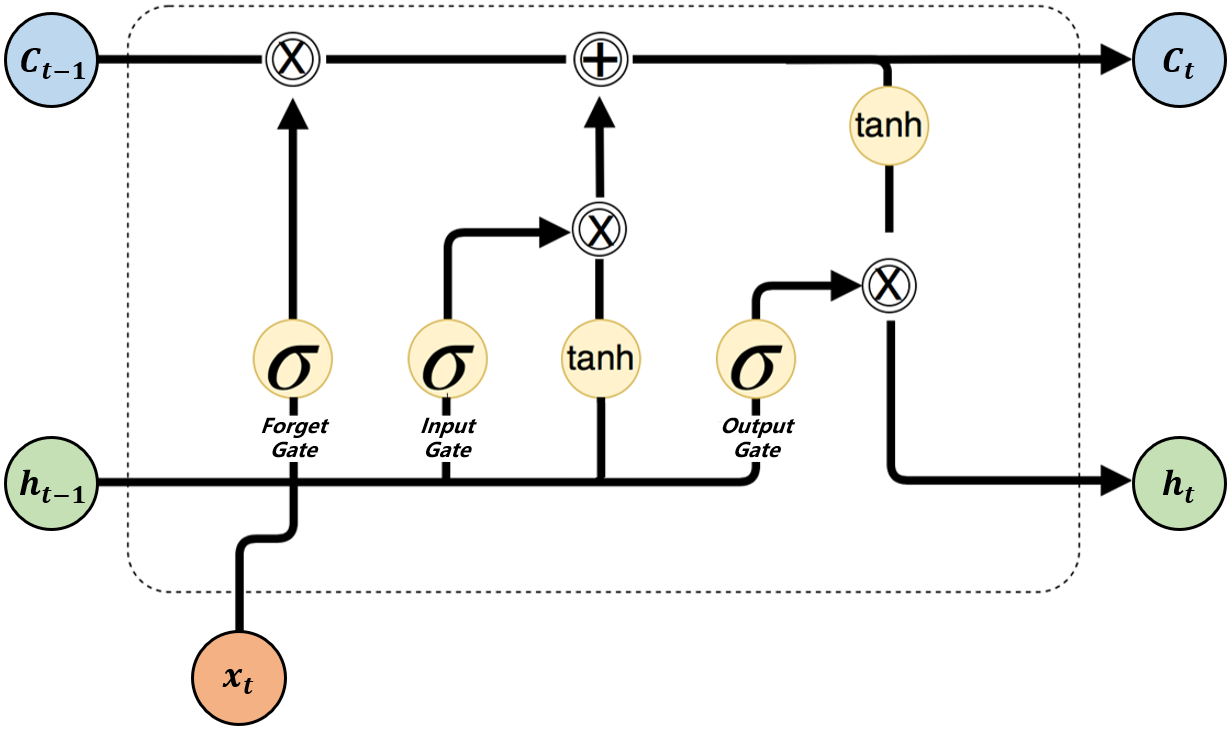}}
\caption{The structure of the LSTM neuron.}
\label{fig:lstm}
\end{figure}

LSTM \cite{hochreiter1997long} is proposed to solve the so-called vanishing/exploding gradient issue. It introduces the concept of “logic gate”, effectively mitigating the problem that RNN is difficult to deal with long-term sequential data. LSTM has three gates and one cell in each hidden neuron (see Figure \ref{fig:lstm}), they are the input gate, the forget gate, the output gate and the memory cell ($c$). Among them, the input gate and the forget gate determine which information ($h_{t-1}, x_{t}$) needs to be remembered and eliminated in the current neuron; the memory cell updates the information in the current neuron: saves the information determined by the input gate to be remembered, and discards the information that is eliminated by the forget gate; and the output gate determines the final information ($h_{t}$) needs to be transmitted to the next neuron. All in all, the retention mechanism for valid information in LSTM not only solves the vanishing/exploding gradient problem that has been plaguing RNN to a great extent, but also helps the network speed up the convergence even if LSTM has four times the amount of parameters of RNN.

However, when the dependencies between the input data are too long, it is still difficult for a vanilla LSTM to accurately grasp the dependencies across time steps. 
Thus, the Attention mechanism \cite{yang2016stacked, seo2016bidirectional} is invented to help neural networks to learn the interconnections between input data. 
As a representative work that promotes the attention mechanism, Transformer \cite{vaswani2017attention} is designed to process sequential data by using a self-attention mechanism that allows it to weigh the importance of different parts of the input sequence when generating outputs. Unlike traditional RNNs, which process sequential data one element at a time, Transformer can process the entire sequence in parallel, making it much faster and more efficient.
BERT \cite{devlin2018bert} is a pre-trained language model composed of multiple bidirectional stacked encoders in Transformer. The model is pre-trained by using a large number of unlabelled texts in an unsupervised manner, and achieves breakthroughs in many downstream tasks, such as sentiment analysis, part-of-speech tagging and name entity recognition. 
GPT-3 \cite{brown2020language} is another well-known pre-trained model that uses multiple stacked Transformer decoders to produce human-like text. The quality of the text generated by GPT-3 is so high that it can be difficult to determine whether or not it is written by a human.

This research also draws on some excellent ideas in the computer vision field. 
As early as the end of the last century, Lecun, Y. et al. proposed LeNet-5 \cite{lecun1998gradient}, which can be seen as the modern structure of CNNs. The network includes the convolution layer, the pooling layer and the fully-connected layer. Weights and biases contained in these layers will be trained and updated with the back-propagation algorithm. However, compared to traditional machine learning algorithms such as SVM, LeNet-5 does not demonstrate superiority in image classification tasks at that time. 
Until AlexNet \cite{krizhevsky2012imagenet} showed great advantages in the ImageNet 2012 competition, a large number of CNN architectures \cite{zeiler2014visualizing, simonyan2014very, szegedy2014scalable} have been invented and proposed, and structures of those models have become deeper and deeper. But deep models cannot consistently make predictions better. 
To mitigate the problem, He, K. et al. \cite{he2015delving} added residual connections to deep convolution networks and achieved a prediction accuracy beyond human level, that is ResNet. The shortcuts in ResNet allow features in shallow layers to be preserved at deeper layers in forward. That means more gradients could impact the shallow layers that are very hard to get updated in normal deep networks. Therefore, ResNet is used as the backbone of many deep models, as is the model proposed in this paper.

%% file: ae_method.tex
\section{Methodology - ResNLS}
\label{sec:method}

\begin{figure}[t]
\centerline{\includegraphics[width=32pc]{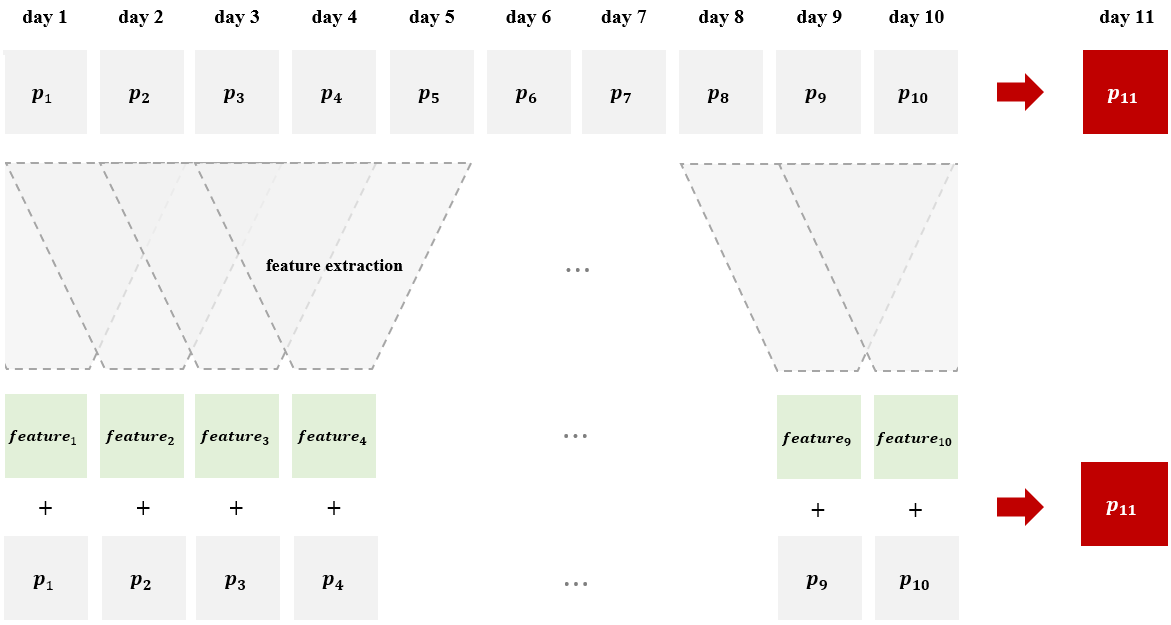}}
\caption{Research ideas of ResNLS.}
\label{fig:idea}
\end{figure}

Most stock price prediction models fail to consider a very important factor that is the dependencies between stock prices across time. Our proposed model, ResNLS, takes into consideration the dependency differences and time-series characteristics between stock prices. For example, suppose we use the stock prices of previous 10 consecutive trading days $ p_{1}, ... , p_{10} $ to predict the stock price of the next trading day $ p_{11} $. When a sequential neural network such as LSTM learns $ p_{10} $, it does not consider the difference between the dependency of $ p_{10} $ and $ p_{9} $, and the dependency of $ p_{10} $ and $ p_{1} $ (see Figure \ref{fig:idea}). However, there should be a stronger relationship between adjacent stock prices than between stock prices farther apart. This is primarily due to the fact that the opening price of the subsequent trading day relies on the closing price of the current day \cite{willinger1999stock}. Consequently, our proposed model focuses on studying the dependency between adjacent stock prices and make predictions based on these dependency features.

\begin{figure}[t]
\centerline{\includegraphics[width=16pc]{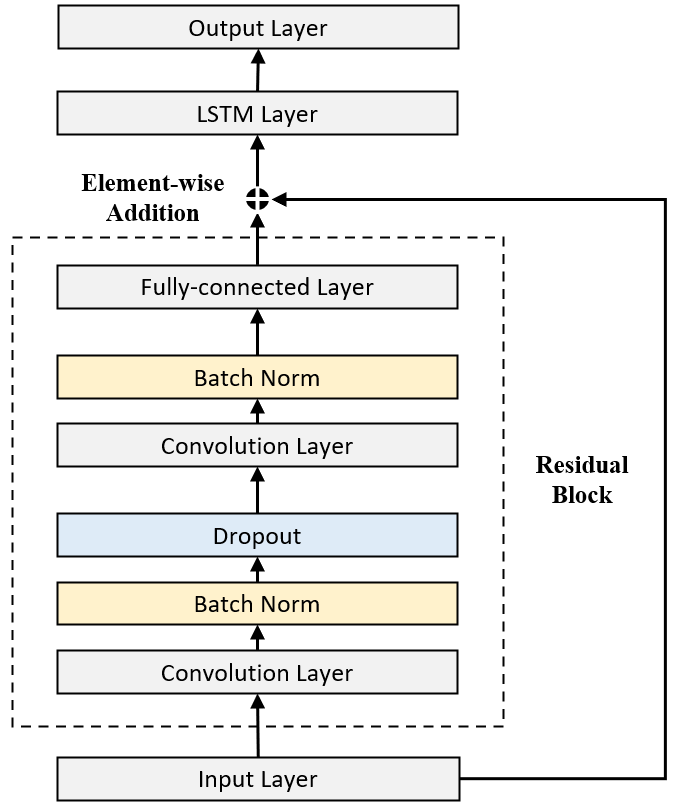}}
\caption{Structure of ResNLS.}
\label{fig:model}
\end{figure}

The structure of our proposed model, ResNLS, is illustrated in Figure \ref{fig:model}. Our model is mainly composed of two kinds of neural networks, ResNet and LSTM, thus named as ResNLS. 
The residual block is the core design, which has a similar structure to the encoder in Transformer. The difference is that the self-attention mechanism in Transformer weighs the influence of all parts of the input sequence on the current input, while the convolution layer in the residual block pays much attention to the influence from adjacent inputs, which will be considered as supplements to the global information.
The residual block is also designed to mitigate the potential problem that weights in shallower layers are hard to be updated during the back-propagation process. Thus, our proposed model can not only draw insights from the original input data, but also learn more data patterns from the dependency features in an effective way.
As for details of the block: first, two convolution layers with the same kernel size are stacked in the block, performing feature extraction on stock price data in sequence. Second, a batch normalization layer and a dropout layer follow behind, aiming to speed up the convergence, as well as making the training process more stable. Third, a fully-connected layer is applied to transform the dimensions of the dependency features so that they will be consistent with the dimensions of the initial input.

Going a step further into the model design, as there are different degrees of dependencies between stock prices as mentioned above, ResNLS uses convolution kernels to pre-process the input stock price data, which is equivalent to feature extraction on stock price data over a fixed time step. 
The feature representations are the vectorized expressions that reflect the dependencies between adjacent stock prices. They will help to deepen the understanding of local dependencies of the input utterance when they provide useful information. On the other hand, they will not have a negative impact on prediction performance if what they represent is not that important. This is because the residual block establishes a shortcut connection between the initial input and the LSTM layer, and gradients from the LSTM layer can be directly transferred to the lower layer, which not only effectively prevents the vanishing gradient problem (the gradients flowing through the network are sometimes too small), but also improves the computational efficiency at the same time.
Afterwards, those features are regarded as a residual term and added element-wise with the initial input to form a new set of sequential data, which is subsequently passed through the LSTM layer on top of the model to make predictions for future stock prices.

%% file: ae_results.tex
\section{Experimental Results}
\label{sec:experiment}

\subsection{Dataset}

The datasets used in this paper are SSE Composite Index and CSI 300 Index (see Table \ref{tab:notations}) obtained from Tushare API \footnote{\url{https://tushare.pro/document/2?doc_id=14}}. The reason why we use them is that the two indices enjoy a high reputation in the global capital market, both of them are known as barometers of China's economy. These two sets contain the stock price data of the indices for each trading day, such as the opening price, the closing price, the highest price as well as the lowest price. Table \ref{tab:sampledata} shows the data sample obtained by segmenting the closing price data of the SSE Composite Index with the 10 consecutive trading days as the unit step. To ensure that the input data is clean, consistent, and has unified structure, we perform data pre-processing on the raw dataset. Particularly, we clean the stock price data, normalize it to eliminate scale differences, and transform it into tensors. We then split the whole pre-processed data into training and test sets. Specifically, we select the closing price data of the SSE Composite Index from January 1, 2011 to December 31, 2020 as the training set, and the closing price data from January 1, 2021 to December 31, 2021 as the test set. Similarly, we select the closing price data of the CSI 300 Index from January 1, 2009 to December 31, 2018 as the training set, while the closing price data from January 1, 2019 to December 31, 2019 as the test set. The selected data intervals enable our model to compare with other baselines.

\begin{table}
\caption{Sample of the pre-processed dataset from the SSE Composite Index.}
\label{tab:sampledata}
\centering
\begin{tabular}{ c c c }
\hline
No. & Input & Target \\
\hline
1 & [0.2806 0.2763 0.2718 0.2763 0.2617 0.2655 0.2709 0.2729 0.2616 0.2353] & 0.2360 \\
2 & [0.2763 0.2718 0.2763 0.2617 0.2655 0.2709 0.2729 0.2616 0.2353 0.2360] & 0.2512 \\
3 & [0.2718 0.2763 0.2617 0.2655 0.2709 0.2729 0.2616 0.2353 0.2360 0.2512] & 0.2262 \\
4 & [0.2763 0.2617 0.2655 0.2709 0.2729 0.2616 0.2353 0.2360 0.2512 0.2262] & 0.2379 \\
5 & [0.2617 0.2655 0.2709 0.2729 0.2616 0.2353 0.2360 0.2512 0.2262 0.2379] & 0.2318 \\
6 & [0.2655 0.2709 0.2729 0.2616 0.2353 0.2360 0.2512 0.2262 0.2379 0.2318]	& 0.2262 \\
7 & [0.2709 0.2729 0.2616 0.2353 0.2360 0.2512 0.2262 0.2379 0.2318 0.2262]	& 0.2359 \\
8 & [0.2729 0.2616 0.2353 0.2360 0.2512 0.2262 0.2379 0.2318 0.2262 0.2359]	& 0.2485 \\
9 & [0.2616 0.2353 0.2360 0.2512 0.2262 0.2379 0.2318 0.2262 0.2359 0.2485]	& 0.2496 \\
10 & [0.2353 0.2360 0.2512 0.2262 0.2379 0.2318 0.2262 0.2359 0.2485 0.2496] & 0.2614 \\
\hline
\end{tabular}
\end{table}

\subsection{Training Procedure}

As for model training, inspired by the self-attention architecture in Transformer, we adopt two convolution layers in the residual block. Each convolution layer uses 64 filters with a kernel size of 3, applies ReLU as the activation function, and adopts a weight decay of 1e-5. The feature data extracted by convolution layers will be normalized by batch normalization layers. Furthermore, a dropout layer with a retain probability of 0.8 is introduced to prevent over-fitting. On top of the residual block, our model applies a LSTM layer whose hidden size is set to 32. 
Furthermore, Adam has been selected as the optimizer due to its ability to implement adaptive learning rates for different parameters in the network, thus a smaller batch size (64) with a larger initial learning rate (1e-3) are preferred \cite{kingma2014adam}. Finally, 50-training-epoch is enough for the model to find the pattern of the stock price data.


\subsection{Evaluation Metrics}

We use Mean Absolute Error (MAE), Mean Squared Error (MSE) and Root Mean Squared Error (RMSE) to evaluate the forecasting performance of models. The smaller the error, the more accurate the prediction. Their equations are shown as follow, $ N_{test} $ is the number of data samples in the test set $ T_{test} $, while $ p_i $ and $ \hat{p_i} $ are the ground truth price and the predicted stock price at the time step $ i $, respectively.

\begin{equation}\label{eq1}
MAE = \frac{1}{N_{test}} \sum_{y \in T_{test}} \vert p_i - \hat{p_i} \vert
\end{equation}

\begin{equation}\label{eq2}
MSE = \frac{1}{N_{test}} \sum_{y \in T_{test}} (p_i - \hat{p_i})^2
\end{equation}

\begin{equation}\label{eq3}
RMSE = \sqrt{\frac{1}{N_{test}} \sum_{y \in T_{test}} (p_i - \hat{p_i})^2}
\end{equation}

We also use Average Rate of Return (ARR) to assess the profitability of trading strategies. The higher the value, the more profitable the strategy is. In the equation below, $ V_0 $ and $ V_n $ represent the asset value at the beginning and end of the strategy period, respectively.

\begin{equation}\label{eq4}
ARR = \frac{V_n -V_0}{V_0} * 100\%
\end{equation}

\subsection{Exploratory Data Analysis}

\begin{table}[b]
\caption{Prediction results for the SSE Composite Index from different inputs.}
\label{tab:inputresults}
\begin{center}
\begin{tabular}{ c c c c c }
\hline
Model & n & MAE & MSE & RMSE \\
\hline
ResNLS-3 & 3 & 32.81 & 1878.67  & 43.34 \\
ResNLS-5 & 5 & \pmb{28.08} & \pmb{1350.16}  & \pmb{36.74} \\
ResNLS-10 & 10 & 38.93 & 2259.17  & 47.53 \\
ResNLS-20 & 20 & 34.32 & 1890.47  & 43.48 \\
ResNLS-40 & 40 & 45.94 & 3074.45  & 55.45 \\
ResNLS-60 & 60 & 68.82 & 6293.34  & 79.33 \\
\hline
\end{tabular}
\end{center}
\end{table}

\begin{figure}
\centerline{\includegraphics[width=32pc]{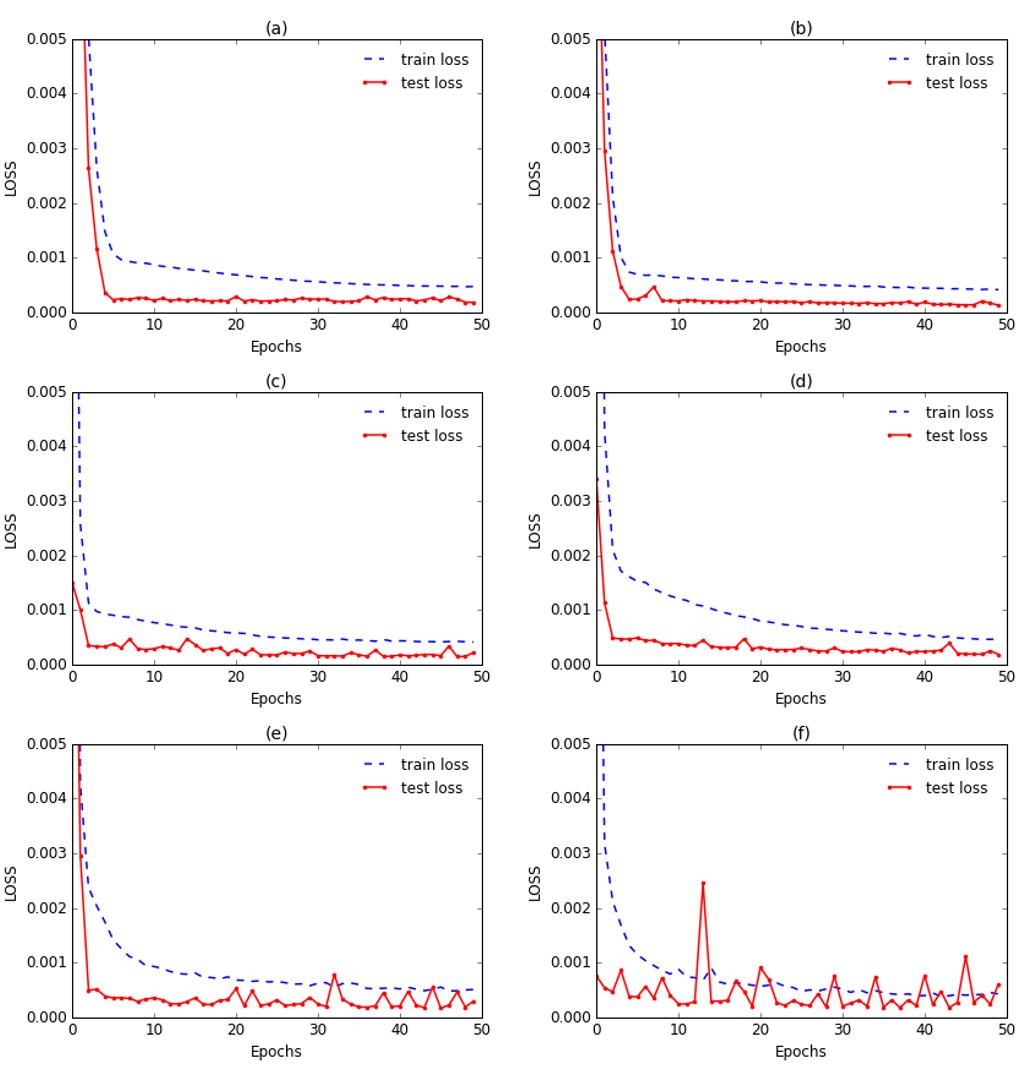}}
\caption{Prediction loss of ResNLSs with different inputs. For each sub-graph: (a) “n” equals to 3; (b) “n” equals to 5; (c) “n” equals to 10; (d) “n” equals to 20; (e) “n” equals to 40; (f) “n” equals to 60.}
\label{fig:inputresults}
\end{figure}

To identify the optimal input data for ResNLS, comparative experiments are conducted in this study. The experiments use the closing price data for the previous "n" (3, 5, 10, 20, 40 and 60) consecutive trading days as input, where those time step sizes are often used in stock analysis \cite{ming2006profitability, huang2017board}. All the experimental models use the same training and test sets of the SSE Composite Index. It is clear from the result (see Table \ref{tab:inputresults}) that ResNLS with the input of the closing price data for the previous 5 consecutive trading days has the smallest prediction errors (MAE, MSE and RMSE), significantly outperforms those models with inputs of closing price data for the previous 3, 10, 20, 40, 60 consecutive trading days.

Figure \ref{fig:inputresults} shows the training loss and test loss of ResNLSs under different inputs. It can be concluded that with the increase of “n”, the fluctuation degree of the test loss becomes more and more serious. When “n” equals to 60, the fluctuation trajectory of the test loss is the most volatile. However, when “n” equals to 5, that is to say, the closing price data for the previous 5 consecutive trading days is used as the input, the fluctuation trajectory of the test loss is the smoothest. This demonstrates that ResNLS-5 can smoothly converge towards the global optimum. In summary, the closing price data for the previous 5 consecutive trading days is the most suitable input for the proposed model, and ResNLS-5 is the best model proposed in this paper.

\subsection{Model Comparison}

Our research also conducts multiple comparative experiments to verify the validity and stability of the best proposed model, ResNLS-5. In addition to ResNLS-5, the experiments also include the vanilla CNN, RNN, LSTM and BiLSTM models, as well as four advanced models including VGG-19 \cite{simonyan2014very}, DenseNet-121 \cite{huang2017densely}, ALSTM \cite{wang2021forecasting} and CNN-BiLSTM-ECA \cite{chen2021stock}. 
It is clear from the experiment results (see Table \ref{tab:modelresults}) that ResNLS-5 has the smallest prediction errors (MAE, MSE, RMSE) in prediction tasks of the SSE Composite Index and CSI 300 Index, significantly outperforms the vanilla CNN, RNN, LSTM, BiLSTM models, and also shows superiority compared to the baselines proposed from recent papers.

\begin{table}
\caption{Prediction results from different models.}
\label{tab:modelresults}
\begin{center}
\begin{tabular}{ c c c c c c c }
\hline
\multirow{2}*{Model} & \multicolumn{3}{c}{SSE Composite Index} & \multicolumn{3}{c}{CSI 300 Index}  \\
~ & MAE & MSE & RMSE & MAE & MSE & RMSE \\
\hline
CNN & 74.61 & 7148.86 & 84.55 & 150.64 & 26872.33 & 163.93 \\
RNN & 67.45 & 5855.22 & 76.52 & 74.95 & 8263.88 & 90.76 \\ 
BiLSTM & 66.92 & 5767.08 & 75.94 & 60.89 & 5827.00 & 76.73 \\ 
VGG-19 & 64.91 & 5521.21 & 74.14 & 61.08 & 5472.30 & 73.97 \\ 
LSTM & 47.57 & 3320.77 & 57.63 & 54.67 & 4923.83 & 70.17 \\
DenseNet-121 & 46.83 & 3049.80 & 55.22 & 57.68 & 54956.15 & 70.30 \\ 
ALSTM  & 28.52 & 2065.25 & 45.45& - & - & - \\
CNN-BiLSTM-ECA & 28.35 & 1956.04 & 44.23 & 39.11 & 3434.41 & 58.60 \\
ResNLS-5 & \pmb{28.08} & \pmb{1350.16} & \pmb{36.74} & \pmb{37.34} & \pmb{2762.82} & \pmb{52.56} \\
\hline
\end{tabular}
\end{center}
\end{table}

\begin{figure}[b]
\centerline{\includegraphics[width=32pc]{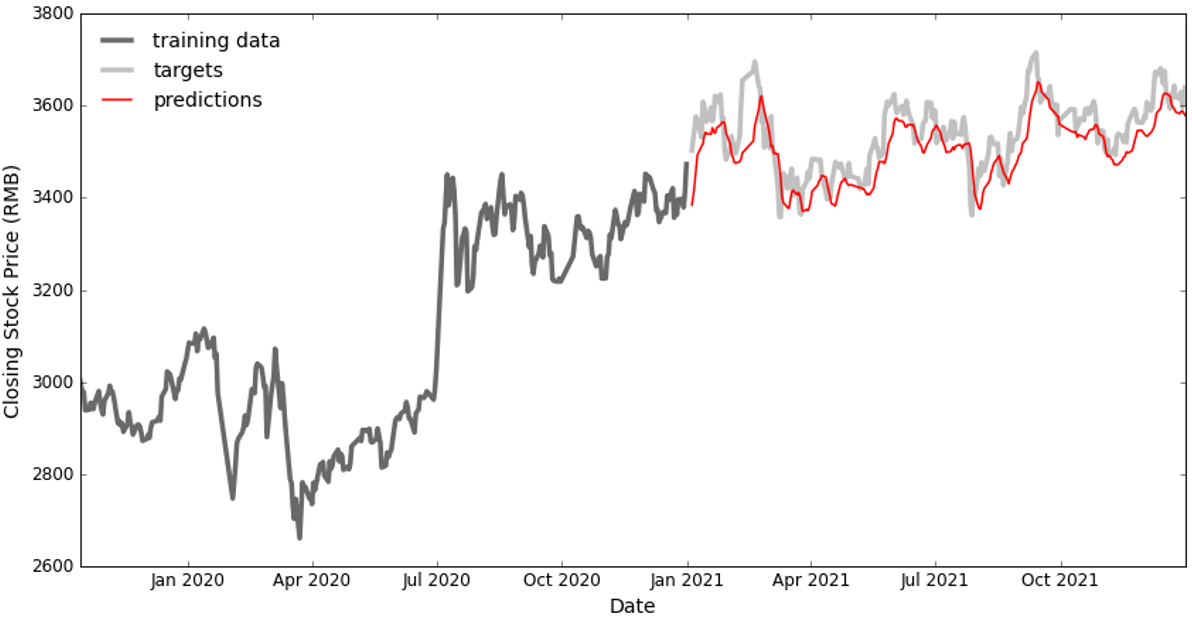}}
\caption{Prediction results of the SSE Composite Index from LSTM.}
\label{fig:lstmforsse}
\end{figure}

\begin{figure}
\centerline{\includegraphics[width=32pc]{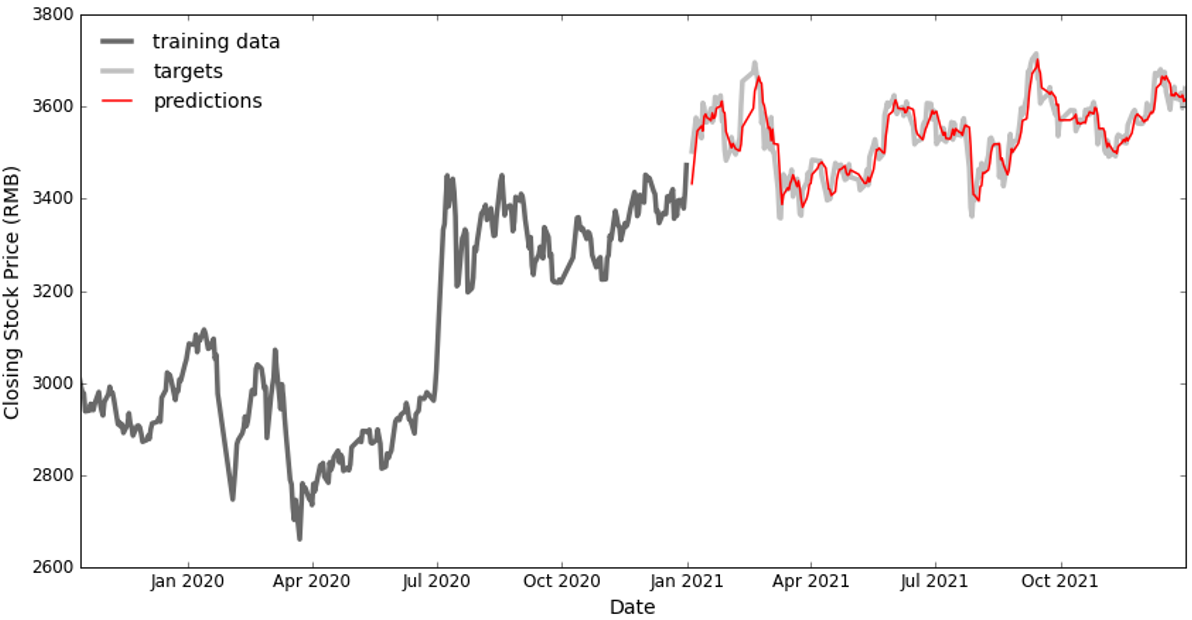}}
\caption{Prediction results of the SSE Composite Index from ResNLS-5.}
\label{fig:resnlsforsse}
\end{figure}

In order to more intuitively compare the prediction performance of the models before and after the improvement, this paper further plots the prediction results of the vanilla LSTM and ResNLS-5 respectively. Figure \ref{fig:lstmforsse} shows the prediction results of the vanilla LSTM on the closing price data of the SSE Composite Index on all trading days from January 1, 2021 to December 31, 2021; Figure \ref{fig:resnlsforsse} shows the prediction results of ResNLS-5 for the closing price data of the SSE Composite Index on all trading days in the same time period. In both figures, the dark grey solid line before 2021 represents the training data. Due to the large amount of data, only part of the training data is shown in the figures; the light grey solid line after 2021 represents the target values in the test data, that is, the real trend of the closing prices of the SSE Composite Index; and the solid red line after 2021 represents the prediction results of the models during this period. Comparing the two figures, it can be seen that most parts of the red solid line in Figure \ref{fig:lstmforsse} are below the light grey solid line, which indicates that there is a significant difference between the prediction results of the vanilla LSTM and the real stock price trend. While the fluctuation trajectories of the red solid line and the light grey solid line in Figure \ref{fig:resnlsforsse} almost coincide, which proves that the prediction performance of ResNLS-5 is more ideal.

\begin{figure}[b]
\centerline{\includegraphics[width=32pc]{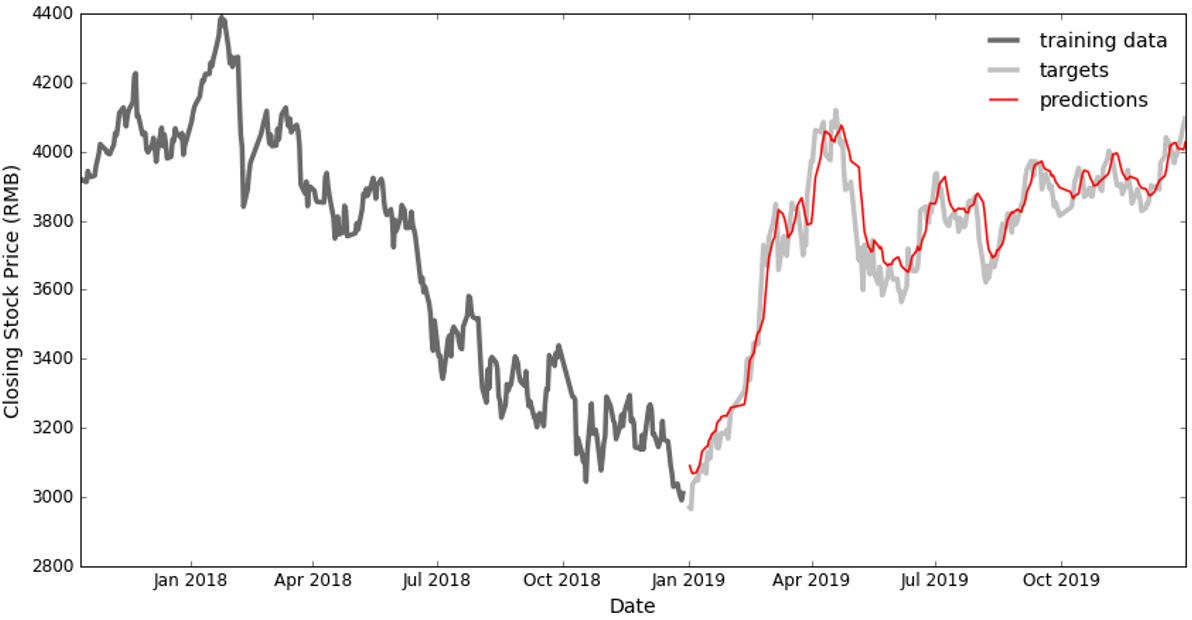}}
\caption{Prediction results of the CSI 300 Index from LSTM.}
\label{fig:lstmforcsi}
\end{figure}

\begin{figure}
\centerline{\includegraphics[width=32pc]{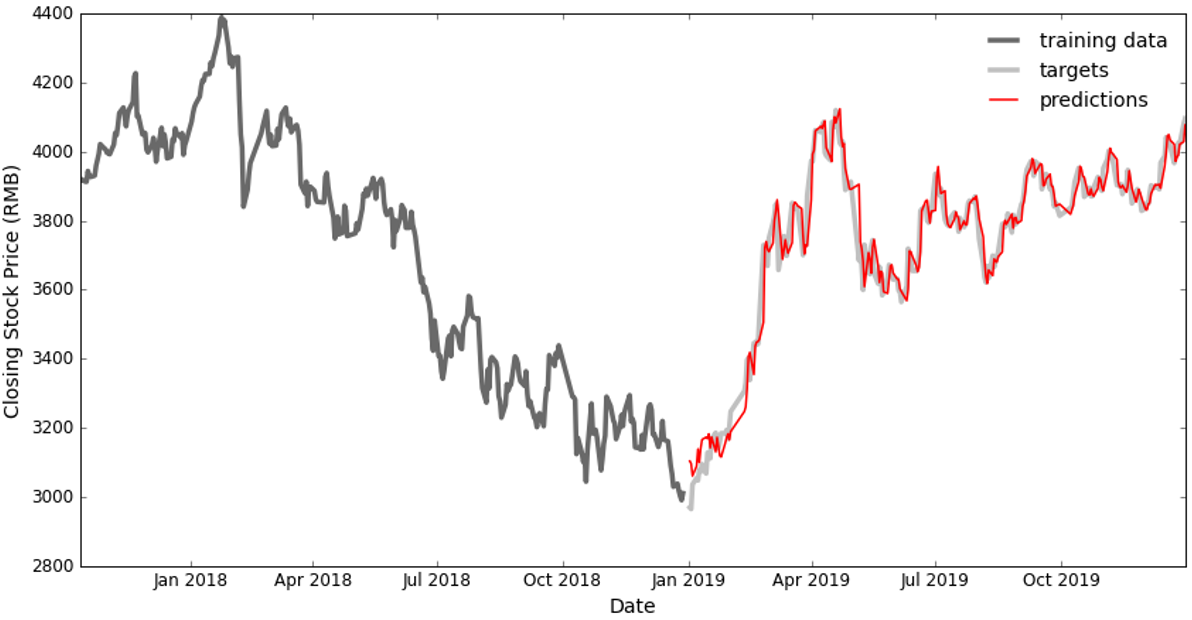}}
\caption{Prediction results of the CSI 300 Index from ResNLS-5.}
\label{fig:resnlsforcsi}
\end{figure}

Similarly, Figure \ref{fig:lstmforcsi} shows the prediction results of the vanilla LSTM on the closing price data of the CSI 300 Index on all trading days from January 1, 2019 to December 31, 2019; Figure \ref{fig:resnlsforcsi} shows the prediction results of ResNLS-5 for the closing price data of the CSI 300 Index on all trading days in the same time period. In both figures, the dark grey solid line before 2019 represents the training data. Due to the large amount of data, only part of the training data is shown in the figures; the light grey solid line after 2019 represents the target values in the test data, that is, the real trend of the closing prices of the CSI 300 Index; and the solid red line after 2019 represents the prediction results of the models during this period. Comparing the two figures, it can be seen that most of the fluctuation trajectories of the red solid line and the light grey solid line in Figure \ref{fig:lstmforcsi} are misaligned, which indicates that there is a significant difference between the prediction results of the vanilla LSTM and the real stock price trend. While the fluctuation trajectories of the red solid line and the light grey solid line in Figure \ref{fig:resnlsforcsi} almost coincide, which proves that the prediction performance of ResNLS-5 is indeed more ideal.

\subsection{Back Testing}

To verify whether the best proposed model, ResNLS-5, can help clients effectively avoid risks and earn profits in the stock market, we constructed a quantitative trading framework to back test the effectiveness of the model. The back testing experiments are conducted on two different data sets, one is the closing price data of the SSE Composite Index from January 1, 2021 to December 31, 2021, and the other is the closing price data of the CSI 300 Index from January 1, 2019 to December 31, 2019.

The experiments compare the ARR of two strategies on these two different data sets. 
One strategy is the benchmark strategy, which involves holding the stock unchanged for a period of time. Therefore, the earnings per share brought by this strategy is equal to the difference between the closing price at the end of the period and the closing price at the beginning of the period. 
Another strategy is the prediction strategy, which is to buy and sell the stock based on the predictions of ResNLS-5. Imagine if you could predict the closing price of a stock in the next trading day, how could you make a profit? A simple way is to buy the stock at the opening price on the next trading day if the future closing price is higher than the current closing price. If the future closing price is lower than the current closing price, then sell the stock at the opening price on the next trading day to avoid risk and ensure returns. Therefore, the rules of the prediction strategy are designed as follows: if the forecast closing price is higher than 1\% of the current closing price, buy all the cash as shares; and if the forecast closing price is lower than 1\% of the current closing price, sell all shares. 
Our experiments do not consider the commissions, taxes and other expenses that may be incurred in the trading process. 
Stock transaction fees vary from place to place, and those expenses have less impact on the final stock return. 
Taxes do have an impact on financial performance, especially for active trading. That is the reason why the threshold is set to 1\%, much higher than the average daily growth rate (0.28\%-0.39\%) of the indexes mentioned above, which also means that the trading conditions are met only 3-4 times during the year and the tax burden will not reverse the conclusion.
All in all, we try to keep the experimental setup as simple and efficient as possible.

\begin{table}[t]
\caption{Back testing results.}
\label{tab:backtestresults}
\begin{center}
\begin{tabular}{ c c c c }
\hline
Data & Time & Strategy & ARR (\%) \\
\hline
SSE Composite Index & 2021-01-01 to 2021-12-31 & Benchmark & 3.91 \\
SSE Composite Index & 2021-01-01 to 2021-12-31 & \pmb{Prediction} & \pmb{11.14} \\
CSI 300 Index & 2019-01-01 to 2019-12-31 & Benchmark & 37.95 \\
CSI 300 Index & 2019-01-01 to 2019-12-31 & \pmb{Prediction} & \pmb{53.15} \\
\hline
\end{tabular}
\end{center}
\end{table}

\begin{figure}[b]
\centerline{\includegraphics[width=32pc]{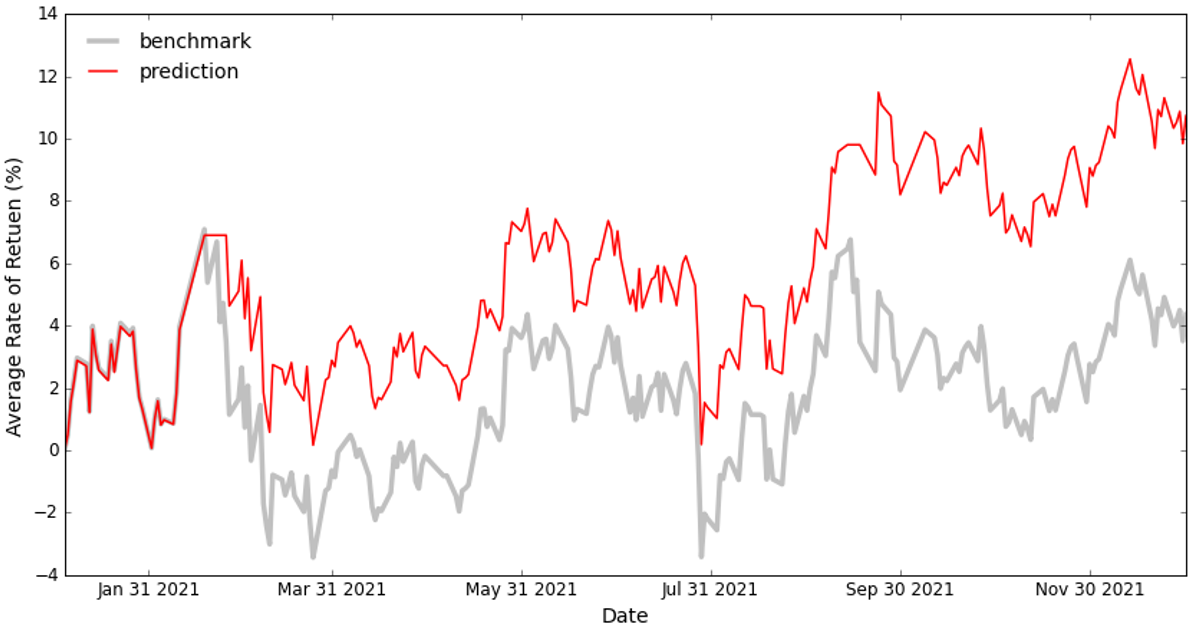}}
\caption{Back testing results of the SSE Composite Index.}
\label{fig:backtestsse}
\end{figure}

\begin{figure}
\centerline{\includegraphics[width=32pc]{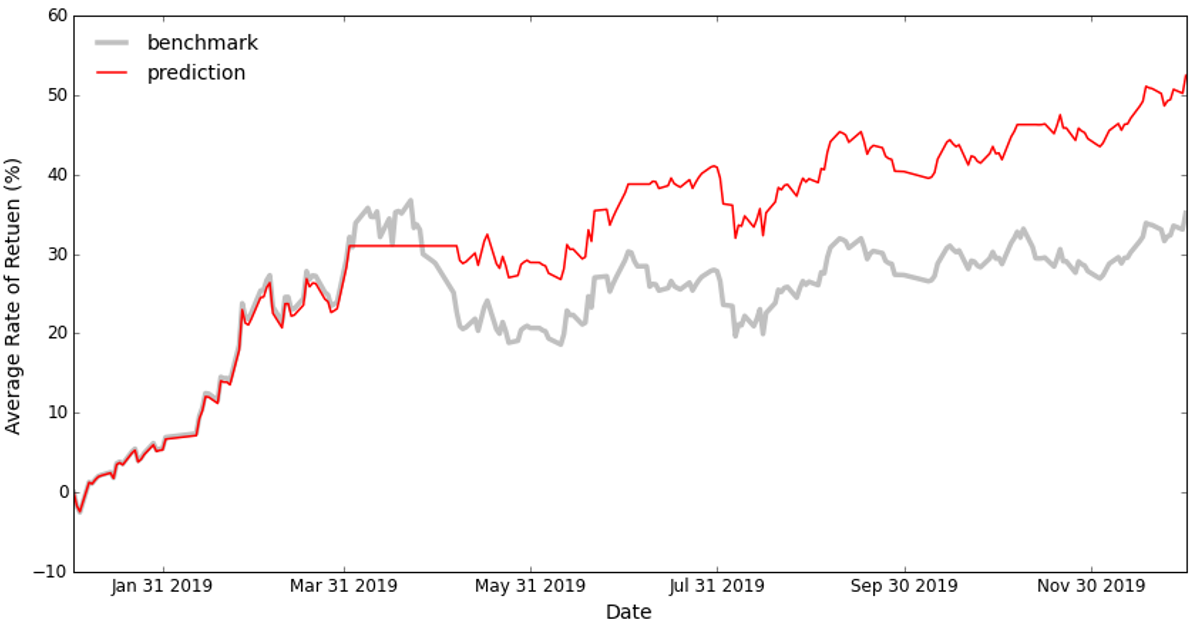}}
\caption{Back testing results of the CSI 300 Index.}
\label{fig:backtestcsi}
\end{figure}

As can be seen from Table \ref{tab:backtestresults}, the ARR of the prediction strategy is higher than that of the benchmark strategy on two different data sets, indicating that applying prediction strategy can bring more profits to the clients. We further plots the back testing results on the two data sets separately. Figure \ref{fig:backtestsse} shows the back testing results on the closing price data of the SSE Composite Index from January 1, 2021 to December 31, 2021; Figure \ref{fig:backtestcsi} shows the back testing results on the closing price data of the CSI 300 Index from January 1, 2019 to December 31, 2019. In both figures, the light grey solid line represents the ARR of the benchmark strategy, and the solid red line represents the ARR of the prediction strategy. Comparing the two lines in each figure, it can be seen that most parts of the red solid line are above the light grey solid line, which proves that the prediction strategy can always bring more profits. We can also find that the prediction strategy effectively avoids the fall of the SSE Composite index at the end of February 2021 (see Figure \ref{fig:backtestsse}), and also avoids the fall of the CSI 300 index in April 2019 (see Figure \ref{fig:backtestcsi}). This avoidance, although its effectiveness may vary during stock declines, still helps clients reduce their exposure to risks.

%% file: ae_conclusion.tex
\section{Conclusion}
\label{sec:conclusion}

The experiments show that the best proposed model in this research, ResNLS-5, has the advantages of high prediction accuracy, high stability and high applicability when predicting future closing stock prices. First, when the closing price data for the previous 5 consecutive trading days is used as the input, the forecast performance of the proposed model is optimal. Furthermore, in prediction tasks of the SSE Composite Index and the CSI 300 Index, ResNLS-5 has the smallest prediction errors, significantly outperforms the vanilla CNN, RNN, LSTM, BiLSTM models, and also shows superiority compared to the advanced models proposed by other papers. Second, the proposed model has strong stability. It can be seen from the above experiments that if the model structure for predicting the SSE Composite Index remains unchanged, and only the training data and test data are replaced with the closing price data of the CSI 300 Index, the model can then be applied to predict the short-term trend of this stock index. Third, the back testing experiments show that ResNLS-5 has high applicability. This is because the prediction strategy based on ResNLS-5 can effectively avoid losses caused by falling stock prices, and at the same time earn profits for clients in the stage of rising stock prices.

The limitations and future work mainly include the following aspects. First, the proposed model in this research only predicts the closing price data. Does the model have the same prediction performance on other stock prices? Future work should include forecasting other stock prices such as the opening price, the lowest price and the highest price of stock indices. Second, the model design is based on the stock price data for the previous 5 consecutive trading days to predict the stock price for the next trading day. While ResNLS-5 does seem to have a better fit, it still appears to have a lag. If one or two months were expanded out for detail, the lag would be even more pronounced. How to predict the long-term stock prices? Future work should consider using macro data to make forecasts of long-term stock price trends. Third, how does the prediction model deal with the impact of unexpected events on stock prices? For example, the recent outbreak of the Russian-Ukrainian war had a disastrous effect on global stock markets. When the black swan event occurs, how to improve the predictions? These are the problems to be solved in the field of stock price forecasting. Therefore, future work should further explore how to handle unexpected events. 

Finally, the authors believe the current AI technology is still in the stage of imitating human capabilities. Although the advantage of computing power shows the potential of surpassing human beings, the AI technology can not perform well when facing with complex tasks that are still difficult for human experts to complete. That is to say, if it is difficult for a stock market specialist with many years of experience to predict long-term price trends and identify unexpected events, we should not expect that the current AI technology will have better performance.